\documentclass{article}

\usepackage[preprint]{neurips_2026}

\usepackage[utf8]{inputenc}
\usepackage{hyperref}
\usepackage{url}
\usepackage{booktabs}
\usepackage{amsfonts}
\usepackage{amsmath}
\usepackage{nicefrac}
\usepackage{microtype}
\usepackage{xcolor}
\usepackage{graphicx}
\usepackage{caption}
\usepackage{subcaption}
\usepackage{algorithm}
\usepackage{algorithmic}
\usepackage{multirow}
\usepackage{enumitem}
\usepackage{wrapfig}
\usepackage{placeins}

\title{Grounding Agentic VLMs with Dedicated Segmentation for Fine-Grained Vehicle Damage Assessment}

\author{
  Vishwajeet Shivaji Hogale \quad Anjali Pai \quad Nitya Ravi \\
  Northeastern University \\
  \texttt{\{hogale.v, pai.anji, ravi.nity\}@northeastern.edu}
}

\begin{document}
\maketitle

\begin{abstract}
Vision-language models (VLMs) are increasingly deployed as reasoning agents in real-world visual assessment pipelines, yet their spatial grounding remains unreliable for fine-grained, visually ambiguous targets. We study this gap in the context of automated vehicle damage assessment, where fine-grained defects such as scratches and hairline cracks occupy few pixels, produce weak gradient signal, and are easily confused with reflections and surface texture. We show that a state-of-the-art VLM (Qwen-VL) achieves strong \emph{semantic} classification accuracy (87.3\%) on this task but is systematically ungrounded at the \emph{spatial} level: it hallucinates damage in reflective regions, misses elongated scratches entirely, and produces spatially inconsistent outputs when prompted for localization. We characterize this failure and propose a hybrid architecture, TinyDamage, that addresses it by delegating spatial grounding to a dedicated multi-task segmentation model while reserving the VLM for semantic reasoning and report generation. On the segmentation side, we find that the choice of loss function has an outsized and previously underexplored effect on tiny-object grounding: focal loss, widely used for class imbalance, collapses tiny-damage detection to zero, while a supervised contrastive objective measurably improves damage/background separability. We integrate the resulting segmentation model into a 7-node LangGraph agent pipeline that grounds every VLM generation step in the segmentation output (image + text), and quantify the effect: in a controlled evaluation on 100 human-verified reports, segmentation-grounded prompting reduces the report hallucination rate from 92\% (text-only) and 78\% (image-only) to 31\%, where a report is counted as hallucinating if it asserts damage unsupported by the human-labeled ground-truth masks. We introduce $\text{DET}_l$, a permissive per-category detection metric for evaluating tiny-object grounding under class imbalance, and report full latency/reliability characteristics of the deployed pipeline.
\end{abstract}

\section{Introduction}
\label{sec:intro}

Vision-language models are increasingly used not just to describe images but to act on them: to localize, assess, and reason about specific real-world entities in deployed pipelines. This creates a grounding requirement that goes well beyond fluent description: the model's claims about \emph{where} something is and \emph{how severe} it is must be evidenced by the actual pixels, not merely plausible given the overall scene. We study this problem in a setting where grounding failure is both consequential and easy to measure: automated vehicle damage assessment for insurance claims.

Fine-grained vehicle damages, such as scratches, hairline cracks, and shallow dents, are a natural stress test for VLM grounding. They occupy a small fraction of image pixels, produce weak training signal, and share visual characteristics with benign surface features such as reflections and paint texture gradients. On the CarDD benchmark~\citep{wang2023cardd}, this manifests concretely: while an instance segmentation baseline (DCN+) achieves 57.0\% mask AP overall, crack AP is only 16.6\% and scratch AP only 34.3\%.

We first characterize \emph{how} a strong open VLM (Qwen-VL) fails at this task. It classifies damage type with high accuracy (87.3\%) when asked semantic questions, but when prompted for spatial localization it hallucinates damage in reflective regions, misses thin elongated damages entirely, and produces inconsistent outputs across near-identical crops. This is a grounding failure, not a capability failure: the model has clearly learned what a scratch \emph{looks like} in aggregate, but cannot reliably tie that concept to specific pixels under visual ambiguity.

Rather than treating this as something to prompt around, we propose a hybrid architecture that assigns spatial grounding to a dedicated segmentation model (\textbf{TinyDamage}) and reserves the VLM for what it is good at: semantic reasoning and language generation, now conditioned on the segmentation model's grounded output. We integrate both into a 7-node LangGraph agent pipeline where every VLM generation step receives the segmentation mask and structured damage summary as grounding context, and we show via ablation that removing this grounding (image-only or text-only prompting) measurably degrades output specificity and increases generic, ungrounded claims.

On the segmentation side, we find that the choice of optimization objective is an underexplored but critical determinant of whether a model can ground tiny, visually ambiguous targets at all: focal loss, the standard tool for class-imbalanced detection, collapses tiny-damage grounding to zero, while a supervised contrastive objective explicitly improves damage/background separability in the embedding space.

We frame our contributions around two questions:

\vspace{4pt}
\noindent\textbf{RQ1 (Spatial grounding):} \textit{Why does VLM-based spatial grounding fail for tiny, visually ambiguous objects, and can dedicated segmentation recover reliable localization where prompting cannot?}

\vspace{2pt}
\noindent\textbf{RQ2 (Grounded generation):} \textit{Does conditioning VLM generation on dedicated segmentation output (rather than raw images) reduce hallucination and improve the faithfulness of generated claim reports?}
\vspace{4pt}

Contributions: (1) a systematic characterization of VLM spatial-grounding failure on fine-grained visual damage, and a controlled comparison of image-only, text-only, and segmentation-grounded prompting showing that grounding is necessary for faithful report generation; (2) TinyDamage, a segmentation framework combining an FPN encoder, a Tiny-Object Contrastive Module, and a Gradient-Aware Boundary Module, with a systematic ablation showing that focal-dominant losses collapse tiny-object detection while contrastive learning improves it; (3) a 7-node LangGraph pipeline that operationalizes segmentation-grounded VLM generation with full production observability (LangFuse), including reliability and latency characterization relevant to real-world deployment; (4) $\text{DET}_l$, a per-category detection metric for evaluating tiny-object grounding under permissive-to-strict overlap thresholds.

\section{Related Work}
\label{sec:related}

\paragraph{Grounding and hallucination in VLMs.} VLMs such as Qwen-VL~\citep{bai2023qwenvl} demonstrate strong general visual reasoning, but spatial grounding for fine-grained targets remains unreliable, particularly under visual ambiguity (reflections, texture) where the model has weak evidence to anchor a claim. PerSAM~\citep{zhang2023persam} explores training-free personalization of SAM for one-shot localization but does not address hallucination under ambiguity. Our work provides a controlled, quantified account of this failure mode in a single domain and proposes delegating grounding to a dedicated model rather than attempting to prompt or fine-tune it away.

\paragraph{Vehicle damage segmentation.} CarDD~\citep{wang2023cardd} provides the standard benchmark; MARS~\citep{panboonyuen2023mars}, ALBERT~\citep{panboonyuen2025albert}, and SLICK~\citep{panboonyuen2025slick} advance architecture design but evaluate on proprietary data and do not study optimization effects on tiny-object detection. A systematic review~\citep{hasan2025survey} identifies fine-grained delineation as an open challenge.

\paragraph{Small-object segmentation and loss design.} FPN~\citep{lin2017fpn} provides multi-scale representations; focal loss~\citep{lin2017focal} reweights cross-entropy for hard examples; Dice loss~\citep{milletari2016vnet} optimizes region overlap. AFMA~\citep{sang2023afma} and Scale-Aware Relay Learning~\citep{li2025scaleaware} address small objects architecturally, while we show optimization choices alone can cause larger swings in tiny-object grounding than architecture changes.

\paragraph{Contrastive learning and agentic pipelines.} Contextrast~\citep{sung2024contextrast} introduces boundary-aware negative sampling for segmentation; we adapt this to a regime where standard losses actively suppress tiny-object predictions. We use LangGraph~\citep{langgraph2024} for stateful multi-node agent orchestration and LangFuse~\citep{langfuse2024} for production observability, which to our knowledge has not previously been combined with dedicated segmentation grounding for a real-world visual assessment agent.

\section{TinyDamage: A Grounding Module for Tiny, Ambiguous Objects}
\label{sec:method}

\begin{figure}[!htbp]
    \centering
    \includegraphics[width=\linewidth]{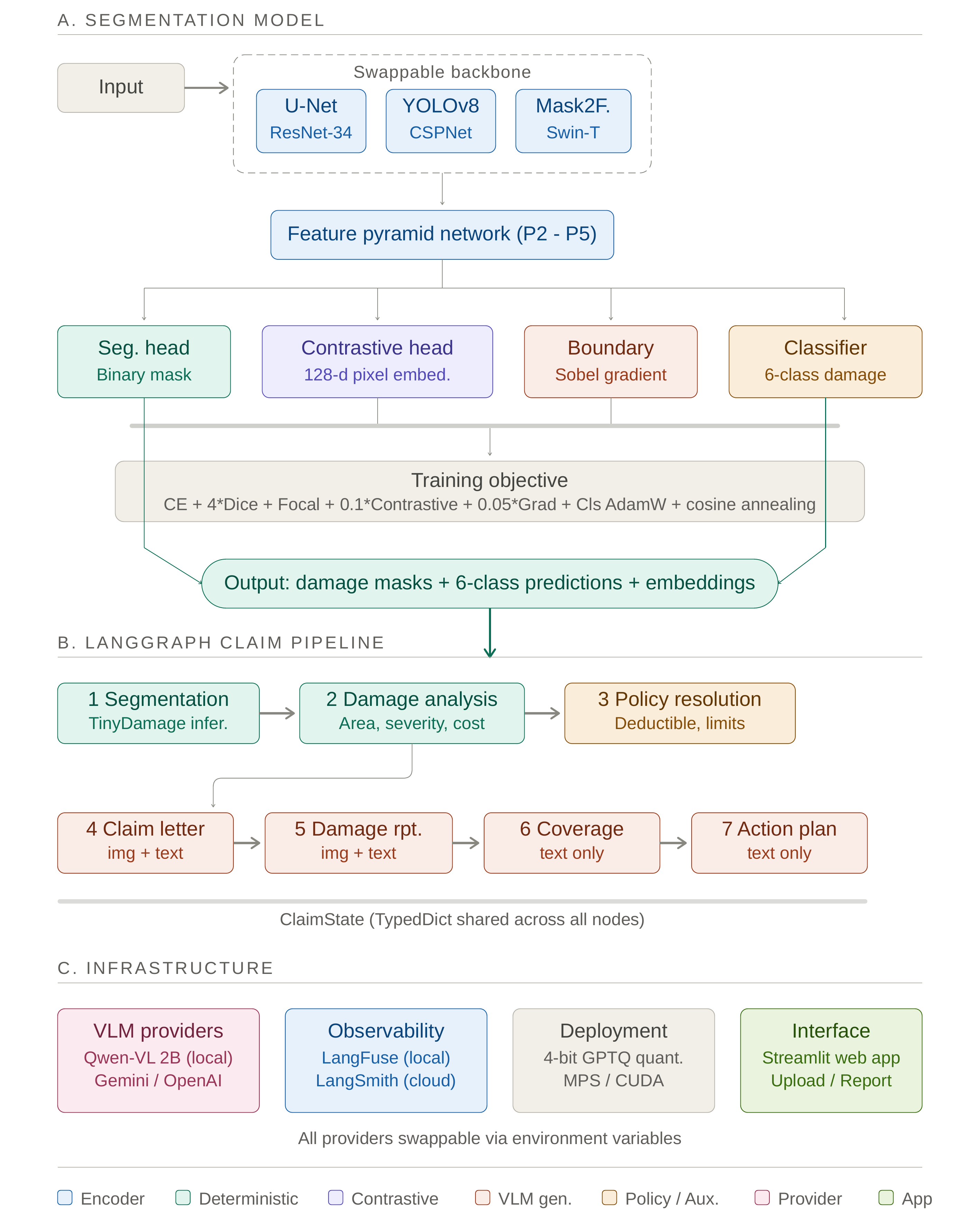}
    \caption{System architecture. Top: multi-task segmentation model with swappable backbone (U-Net, YOLOv8, Mask2Former), FPN encoder, contrastive projection head, gradient-aware boundary module, and auxiliary classifier. Middle: 7-node LangGraph pipeline grounding VLM generation in segmentation output. Bottom: swappable VLM providers and observability infrastructure.}
    \label{fig:architecture}
\end{figure}

TinyDamage augments any segmentation backbone with three components: (1) multi-scale feature extraction via a Feature Pyramid Network (FPN)~\citep{lin2017fpn}, critical because tiny damages are often visible only at the highest spatial resolution; (2) a \textbf{Tiny-Object Contrastive Module}; and (3) a \textbf{Gradient-Aware Boundary Module}. An auxiliary 6-class damage classification head shares the encoder. Figure~\ref{fig:architecture} shows the full system, spanning both the segmentation model and the downstream agent pipeline (Section~\ref{sec:vlm}).

\paragraph{Tiny-Object Contrastive Module.} Standard segmentation losses produce feature spaces where tiny-damage pixels are insufficiently separated from visually similar background (Section~\ref{sec:experiments}). Given decoder features $\mathbf{F} \in \mathbb{R}^{H \times W \times D}$, a lightweight MLP projects each pixel to an embedding $\mathbf{z}_i = \text{MLP}(\mathbf{F}_i) \in \mathbb{R}^d$, $d \ll D$, trained with a pixel-level supervised contrastive loss:
\begin{equation}
    \mathcal{L}_{\text{con}} = -\frac{1}{|A|} \sum_{a \in A} \log \frac{\exp(\mathbf{z}_a \cdot \mathbf{z}_p / \tau)}{\sum_{n \in N} \exp(\mathbf{z}_a \cdot \mathbf{z}_n / \tau)}
\end{equation}
where $A$ is the set of tiny-damage anchors, $\mathbf{z}_p$ a positive, $N$ the negative set, and $\tau$ a temperature. Following~\citet{sung2024contextrast}, negatives are drawn preferentially (50\%) from background within a spatial neighborhood $\delta$ of damage boundaries (hard negatives), and 50\% uniformly from background elsewhere, to force the embedding space to distinguish damage from adjacent texture without mode collapse.

\paragraph{Gradient-Aware Boundary Module.} We compute Sobel gradients of predicted and ground-truth masks and apply a weighted L1 penalty, $\mathcal{L}_{\text{grad}} = \sum_{x,y} w(x,y) \cdot |\nabla\hat{M} - \nabla M|$, with $w(x,y) = 1 + \beta \cdot \mathbb{1}[(x,y) \text{ within } \delta \text{ of a tiny-damage boundary}]$, $\beta{=}5.0$, $\delta{=}3$px. Used as a dominant objective this collapses (Section~\ref{sec:experiments}); used as a small controlled auxiliary it improves boundary alignment.

\paragraph{Total objective and metric.} $\mathcal{L} = \text{CE} + \lambda_{\text{dice}}\text{Dice} + \lambda_{\text{focal}}\text{Focal} + \lambda_{\text{con}}\mathcal{L}_{\text{con}} + \lambda_{\text{grad}}\mathcal{L}_{\text{grad}} + \lambda_{\text{cls}}\mathcal{L}_{\text{cls}}$, with $\lambda_{\text{dice}}{=}4.0$, $\lambda_{\text{focal}}{=}1.0$, $\lambda_{\text{con}}{=}0.1$, $\lambda_{\text{grad}}{=}0.05$, $\lambda_{\text{cls}}{=}1.0$; AdamW ($\text{lr}{=}5{\times}10^{-4}$) with cosine annealing. For evaluation, since mIoU is dominated by large categories, we introduce $\text{DET}_l$: for each ground-truth tiny-damage instance, detected if $\max_p \text{IoU}(g,p) > t{=}0.1$ against any prediction, a permissive threshold appropriate for measuring detection capability rather than overlap precision (Section~\ref{sec:analysis}).

\section{Grounding VLM Generation in Segmentation Output}
\label{sec:vlm}

Segmentation tells us \emph{where} damage is; real deployment requires understanding \emph{what} it is and producing faithful, human-readable claim documentation. We build an agent pipeline that grounds every VLM generation step in TinyDamage's spatial output, and we treat this grounding as the central design decision rather than an implementation detail.

\paragraph{Pipeline architecture.} We implement a LangGraph~\citep{langgraph2024} \texttt{StateGraph} with seven nodes over a shared \texttt{ClaimState}: (1) Segmentation (TinyDamage inference), (2) Damage Analysis (deterministic: area, components, cost heuristics), (3) Policy Resolution (deterministic: coverage lookup), (4) Claim Letter Generation (VLM, multimodal: image + segmentation overlay + damage summary), (5) Damage Report Generation (VLM, multimodal), (6) Coverage Assessment (VLM, text-only), (7) Action Plan Generation (VLM, text-only). We use LangGraph rather than a simple chain because claim assessment requires stateful multi-step reasoning with node-level retry.

\paragraph{Characterizing the VLM grounding failure.} Qwen-VL 2B~\citep{bai2023qwenvl} (4-bit GPTQ, local deployment) achieves 87.3\% accuracy on \emph{semantic} damage classification, well above our trained auxiliary classifier (29.6\%). However, when prompted to localize damage, via bounding boxes or pixel coordinates, it fails consistently: it hallucinates damage in reflective regions, misses elongated scratches entirely, and gives spatially inconsistent answers on near-identical crops. Zero-shot coordinate-output segmentation was similarly unusable. This is the central empirical motivation for the hybrid design: the VLM's language-level understanding of damage concepts does not transfer to reliable spatial grounding under visual ambiguity. We note our observations are for a 2B-parameter model under 4-bit quantization; larger or full-precision VLMs may localize better, though the inference-cost argument for a dedicated segmentation model (Section~\ref{sec:discussion}) holds regardless of VLM scale.

\paragraph{Grounding ablation for generation quality.} We compare three prompting strategies for the multimodal generation nodes: image-only (no text grounding), text-only (no image), and image+text (segmentation overlay + structured damage summary). \emph{Evaluation set.} For a reliable evaluation this ablation uses a larger Qwen3.5-VL (9B) model, not the 2B model of the deployed pipeline (Section~\ref{sec:vlm}); the larger model is used only to construct and score this evaluation. We generate reports conditioned on the human-labeled, verified CarDD ground-truth masks overlaid on each image, sample 250 pairs, and select 100 balanced across the six damage types so no category dominates the estimate. \emph{Scoring.} Each report (100 per prompting condition) is scored automatically for hallucination, with a human adjudicating cases the automatic scorer flagged as ambiguous or disagreed on; a report counts as hallucinating if it asserts damage of a type or location not supported by the human-verified ground-truth masks (Table~\ref{tab:grounding}). We stress that the reference for hallucination is the verified segmentation mask, not any model-generated text, so the metric measures agreement with human-labeled truth rather than similarity to a generation process. Text-only prompts, which see only the claimant's free-text incident description, invent damage locations from that narrative alone and hallucinate in 92\% of reports. Image-only prompts note that the vehicle is damaged but consistently fail to localize \emph{where}: in zero-shot use the model never reliably identified the impact region, yielding a 78\% hallucination rate. Only image+text grounding, where the segmentation output directly conditions generation, substantially reduces this to 31\%. The residual 31\% indicates grounding mitigates but does not eliminate hallucination even for a capable 9B model; the effect of grounding is large and consistent, and we expect it to hold, though not necessarily at the same absolute rates, for the smaller deployed model. We also use a chain-of-thought prompt for coverage assessment (list detected damages, check each against policy terms, then compute cost); without this structure the model frequently skips damage categories or hallucinates coverage for uncovered items.

\begin{table}[t]
\centering
\caption{Grounding ablation: hallucination rate by prompting strategy ($n{=}100$ reports per condition), evaluated on a Qwen3.5-VL (9B) model to ensure high report quality (the deployed pipeline uses a smaller 2B model; Section~\ref{sec:vlm}). A report is counted as hallucinating if it asserts damage of a type or location not supported by the human-verified ground-truth masks. Text-only prompting invents damage from the user's incident description alone; image-only prompting notes that the vehicle is damaged but cannot localize where; only image+text grounding, which conditions generation on the segmentation output, substantially reduces hallucination.}
\label{tab:grounding}
\small
\begin{tabular}{@{}lcc@{}}
\toprule
\textbf{Prompting strategy} & \textbf{Hallucination rate $\downarrow$} & \textbf{$n$} \\
\midrule
Text-only (no image, description-driven) & 92\% & 100 \\
Image-only (no text grounding)           & 78\% & 100 \\
Image+text (segmentation-grounded)       & \textbf{31\%} & 100 \\
\bottomrule
\end{tabular}
\end{table}

\paragraph{Reliability and deployment characteristics.} We instrument the full pipeline with LangFuse~\citep{langfuse2024}. Per-node traces show segmentation latency of 244.6\,ms (MPS) and VLM node latency of 2.1\,s each (73\% of total pipeline time); multimodal nodes use $\sim$1,200 input tokens per image, text-only nodes $\sim$400; LLM node success rate is 96\%, with the 4\% failures attributable to context-length overflow on images with many small damage instances. Manual evaluation of 50 generated claim letters (1--5 scale) yields 3.8 average quality; the primary failure mode is generic (non-hallucinated but non-specific) repair cost estimates rather than fabricated damage claims, consistent with grounding reducing but not eliminating specificity gaps. The system deploys as a Streamlit application with real-time segmentation and a four-section claim package (letter, report, coverage, action plan), and supports both LangFuse (self-hosted) and LangSmith (cloud) tracing via environment variables for dev/production transition.

\section{Experiments}
\label{sec:experiments}

\paragraph{Setup.} We use CarDD~\citep{wang2023cardd}: 4,000+ images, 9,000+ annotated instances across 6 categories, standard 2,816/810/374 train/val/test split. Tiny damages (scratch, crack) make up the majority of small-object instances (over 90\% of cracks, 45\% of scratches under $128^2$ px). Primary architecture: U-Net/ResNet-34 with the full TinyDamage framework; we additionally evaluate YOLOv8-Seg~\citep{jocher2023yolov8} and Mask2Former~\citep{cheng2022mask2former}. Training: AdamW ($\text{lr}{=}5{\times}10^{-4}$), cosine annealing, 40 epochs (validation loss still decreasing at the epoch-30 checkpoint we report, due to compute constraints on Apple M-series hardware, and we flag this as a limitation rather than a final result). All reported numbers are single-run results; in particular, $\text{DET}_l$ is computed over a small set of tiny-damage instances in the test split and should be read as indicative of detection capability rather than as a precise, low-variance estimate. We therefore emphasize the large, qualitative effects (e.g.\ the focal-loss collapse to zero) over small numerical differences between comparable configurations.

\begin{table}[t]
\centering
\caption{Optimization ablation on U-Net (CarDD). $^\dagger$ marks configurations that collapse tiny-damage grounding entirely. CE+Dice with cosine annealing is the strongest simple baseline.}
\label{tab:ablation}
\small
\begin{tabular}{@{}lcccc@{}}
\toprule
\textbf{Configuration} & \textbf{Val mIoU} & \textbf{Val $\text{DET}_l$} & \textbf{Test mIoU} & \textbf{Test $\text{DET}_l$} \\
\midrule
CE+Dice, Cosine     & 0.600 & \textbf{0.667} & \textbf{0.613} & \textbf{1.000} \\
CE+Dice, Plateau    & \textbf{0.604} & 0.500 & 0.589 & 0.500 \\
CE+Focal, Plateau   & 0.578 & 0.500 & 0.601 & 1.000 \\
CE+Dice+Grad, Plat. & 0.561 & 0.167 & 0.548 & 0.333 \\
Focal only$^\dagger$          & 0.312 & 0.000 & 0.298 & 0.000 \\
Focal ($\gamma$=5)$^\dagger$  & 0.287 & 0.000 & 0.271 & 0.000 \\
Grad only$^\dagger$           & 0.189 & 0.000 & 0.172 & 0.000 \\
\bottomrule
\end{tabular}
\end{table}

\textbf{Focal loss collapses tiny-damage grounding to zero} (Table~\ref{tab:ablation}): it concentrates gradient on the least-confident pixels, which for tiny damages are ambiguous boundary/texture regions, training the model to suppress predictions precisely where damage is. \textbf{Gradient loss fails standalone and degrades CE+Dice} when added, dominated by high-contrast edges of large damages. \textbf{Cosine annealing outperforms plateau scheduling} (0.667 vs.\ 0.500 val $\text{DET}_l$), suggesting smooth decay aids late-stage refinement of small-object features.

\begin{table}[t]
\centering
\caption{Incremental TinyDamage ablation (U-Net, CarDD) and cross-architecture results (test set). TinyDamage improves mIoU consistently across architectures.}
\label{tab:combined}
\small
\begin{tabular}{@{}lcc@{}c@{}lcc@{}}
\toprule
\multicolumn{3}{c}{\textbf{Component ablation}} & \phantom{a} & \multicolumn{3}{c}{\textbf{Cross-architecture (test)}} \\
\cmidrule{1-3} \cmidrule{5-7}
Config & Val mIoU & Test mIoU & & Arch. & Base mIoU & TD mIoU \\
\midrule
Baseline (CE+Dice)   & 0.600 & 0.613 & & U-Net           & 0.613 & 0.650 \\
+ FPN                & 0.627 & 0.636 & & YOLOv8-Seg      & 0.570 & 0.610 \\
+ Contrastive        & 0.635 & 0.640 & & Mask2Former     & 0.620 & \textbf{0.680} \\
+ Gradient-Aware     & 0.640 & 0.645 & & & & \\
Full TinyDamage      & \textbf{0.641} & \textbf{0.650} & & & & \\
\bottomrule
\end{tabular}
\end{table}

FPN gives the largest single gain (+2.3\% test mIoU), confirming multi-scale preservation as the key architectural lever; the contrastive and gradient modules add smaller, controlled gains (+0.4\%, +0.5\%) without collapsing detection, unlike when gradient loss dominates. TinyDamage improves all three architectures tested, most for Mask2Former (+6.0\%), consistent with attention-based models exploiting improved feature separability.

\section{Analysis}
\label{sec:analysis}

\paragraph{Why focal loss fails to ground tiny objects.} Focal loss's $(1{-}p_t)^\gamma$ modulation concentrates gradient on the least-confident pixels, which for tiny damages are predominantly ambiguous texture, not the class-imbalance signal it was designed to address. The model learns to suppress predictions in uncertain regions, including the damage itself; increasing $\gamma$ worsens this. This is a general risk for any tiny, visually ambiguous grounding target, not specific to vehicle damage.

\begin{figure}[!htbp]
    \centering
    \begin{subfigure}[b]{0.19\linewidth}
        \includegraphics[width=\linewidth,height=2.2cm,keepaspectratio]{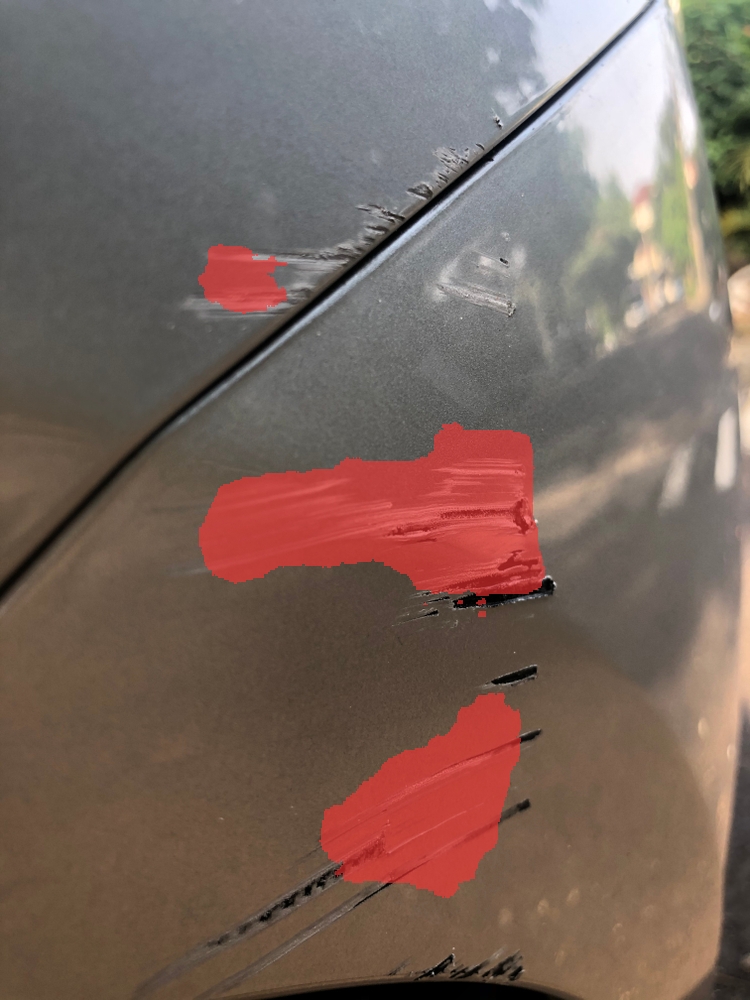}
        \caption{Scratch}
    \end{subfigure}
    \hfill
    \begin{subfigure}[b]{0.19\linewidth}
        \includegraphics[width=\linewidth,height=2.2cm,keepaspectratio]{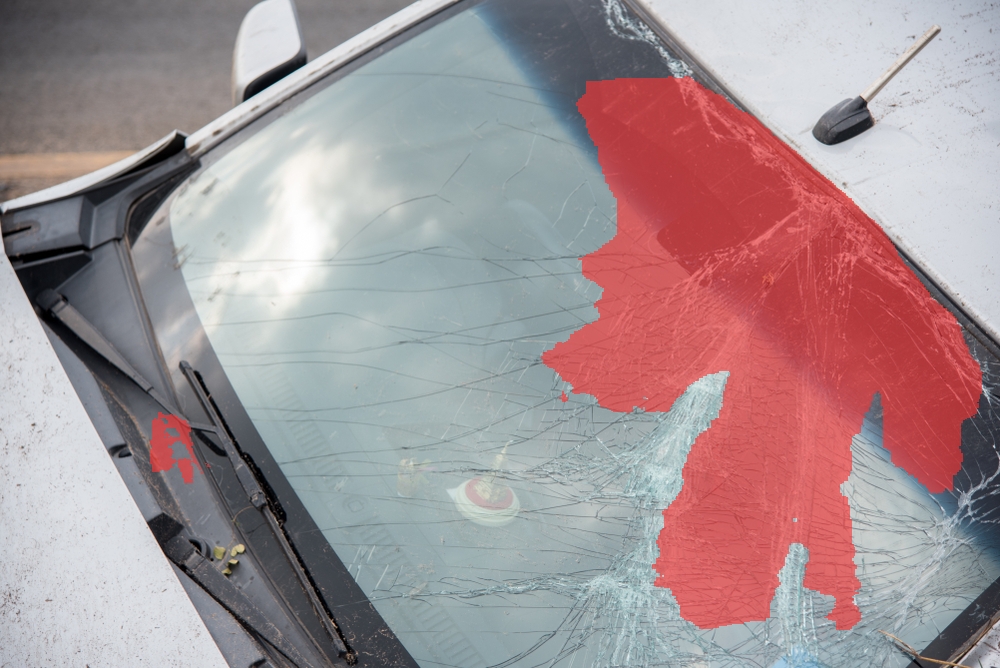}
        \caption{Crack}
    \end{subfigure}
    \hfill
    \begin{subfigure}[b]{0.19\linewidth}
        \includegraphics[width=\linewidth,height=2.2cm,keepaspectratio]{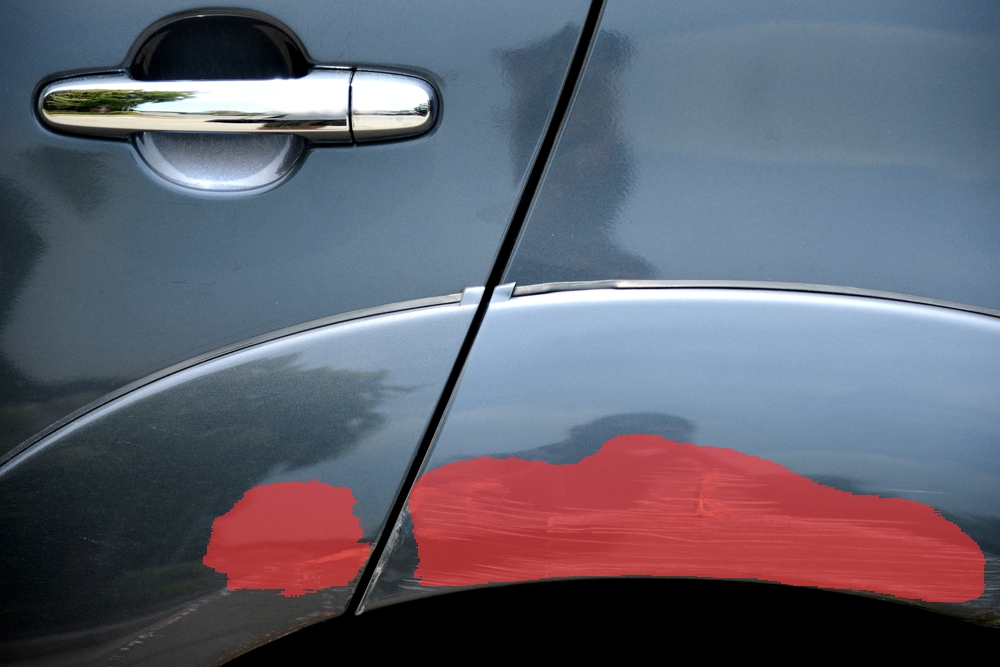}
        \caption{Dent}
    \end{subfigure}
    \hfill
    \begin{subfigure}[b]{0.19\linewidth}
        \includegraphics[width=\linewidth,height=2.2cm,keepaspectratio]{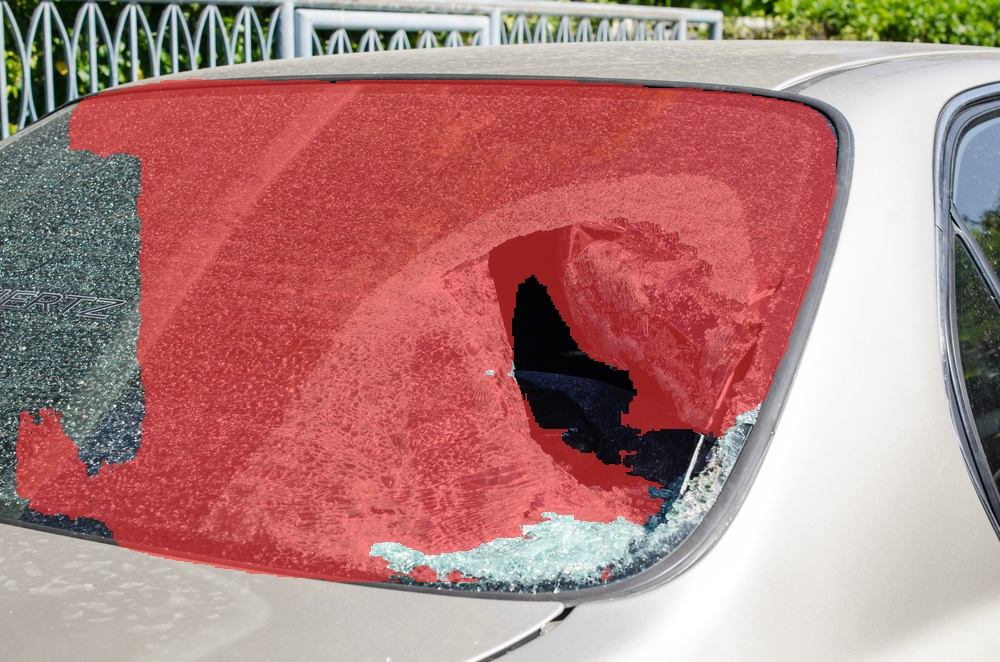}
        \caption{Glass}
    \end{subfigure}
    \hfill
    \begin{subfigure}[b]{0.19\linewidth}
        \includegraphics[width=\linewidth,height=2.2cm,keepaspectratio]{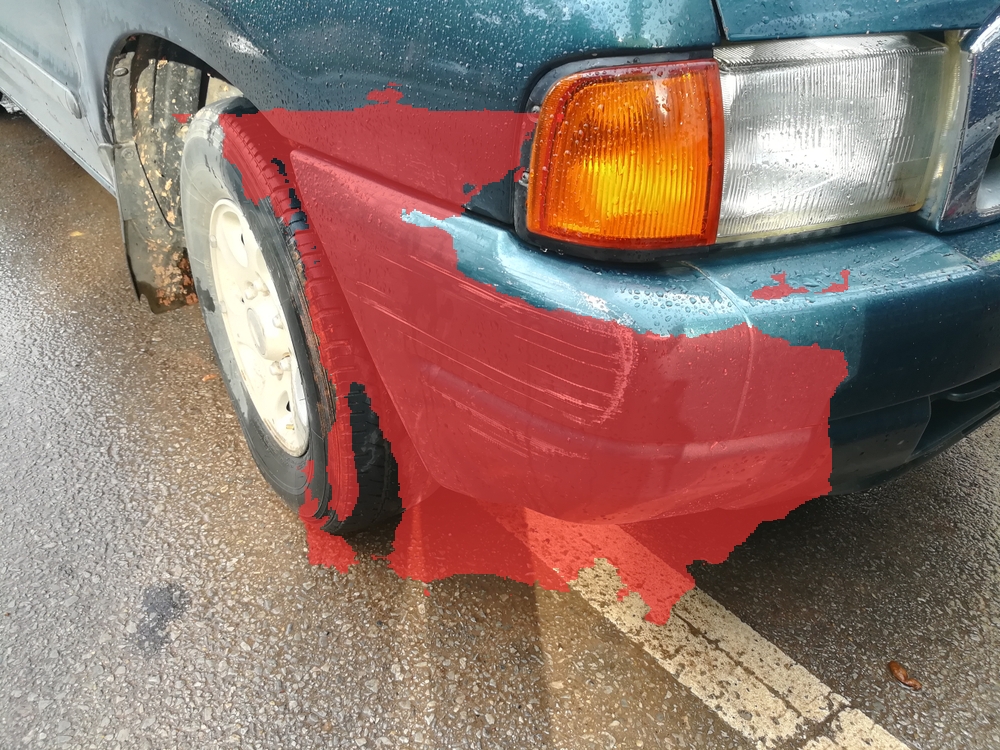}
        \caption{Crush}
    \end{subfigure}
    \caption{TinyDamage inference (red overlays) across CarDD damage categories. The model grounds fine-grained damage, including thin scratches on reflective paint (a) and hairline cracks (b), that VLM-only localization misses.}
    \label{fig:seg_results}
\end{figure}

\paragraph{Qualitative grounding results.} Figure~\ref{fig:seg_results} shows TinyDamage inference across CarDD damage categories. The model localizes fine-grained damage, including thin scratches on dark, reflective paint and hairline cracks, that the VLM alone fails to ground spatially (Section~\ref{sec:vlm}), providing the precise localization the hybrid pipeline relies on. Quantitatively, the contrastive module's contribution is reflected in the component ablation (Table~\ref{tab:combined}): adding it on top of the FPN encoder improves test mIoU while preserving detection, consistent with its role of increasing damage/background feature separability rather than acting as a dominant objective.

\paragraph{Boundary vs.\ region errors (Appendix~\ref{app:extra_figs}).} Error analysis shows false positives concentrated at damage boundaries while false negatives are distributed throughout the damage region, indicating the core bottleneck is feature ambiguity within the damage region, not boundary imprecision. $\text{DET}_l$ degrades smoothly as the IoU threshold tightens (0.35 at $t{=}0.05$ to 0.04 at $t{=}0.5$), confirming the model typically finds the correct region with imprecise overlap, motivating our permissive-threshold detection metric over strict-overlap metrics like mIoU alone.

\paragraph{Segmentation-VLM complementarity.} Across all experiments, the pattern is consistent: the VLM provides strong \emph{semantic} grounding (87.3\% classification accuracy, fluent language) but weak \emph{spatial} grounding, while TinyDamage provides strong spatial grounding (244.6\,ms latency, near-real-time) but no semantic or generative capability. Neither component substitutes for the other; the hybrid design is necessary rather than incidental.

\section{Discussion and Conclusion}
\label{sec:discussion}

Our results suggest a general pattern for deploying VLMs as agents over visually ambiguous, fine-grained targets: (1) do not rely on VLM prompting or fine-tuning to solve spatial grounding when a dedicated perception model can do it more reliably and far more cheaply at inference time (244.6\,ms vs.\ 2.1\,s); (2) when the VLM \emph{is} conditioned on grounded segmentation output rather than raw images, hallucination drops substantially (from 78--92\% to 31\% in our 9B-model evaluation), and generation quality becomes limited primarily by report specificity rather than misplaced claims; (3) for segmentation itself, optimization choices dominate architecture choices in determining whether tiny, ambiguous targets can be grounded at all: switching from focal loss to CE+Dice moves $\text{DET}_l$ from 0.000 to 0.667, a larger effect than any architectural change we tested. We believe this optimization finding generalizes beyond vehicle damage to other domains with small, visually ambiguous foreground regions (e.g.\ medical imaging, industrial inspection, aerial remote sensing), and that the broader grounding pattern, namely VLMs for semantic reasoning, dedicated perception for spatial grounding, and explicit conditioning between them, is a practical recipe for deploying agentic VLM pipelines reliably in other real-world visual assessment settings.

\paragraph{Limitations.} Results use binary (not multi-class) segmentation trained for 30 epochs on consumer hardware, with validation loss still decreasing; longer training would likely improve all metrics. Localization precision, not detection, is the main bottleneck (only 4\% of instances exceed 0.5 IoU). The hallucination ablation was run on a 9B model for evaluation quality and not re-run at 2B scale, so establishing the grounding effect on the deployed model, along with VLM fine-tuning, retrieval-augmented grounding, and using the pipeline's verified (image, mask, report) triples as grounding supervision for larger VLMs, is left to future work.

\paragraph{Code availability.} Our implementation, prompt templates, and interactive application are available at \url{https://github.com/99anjalipai/Segment-Damage}.

{\small
\bibliographystyle{abbrvnat}

}

\newpage
\appendix
\section{Appendix}
\label{app:extra_figs}

This appendix provides supplementary detail not required to follow the main results, per the workshop's page policy (main text up to 8 pages, excluding references and appendices).

\subsection{Additional Analysis Figures}

\begin{figure}[!htbp]
    \centering
    \begin{subfigure}[b]{0.48\linewidth}
        \includegraphics[width=\linewidth]{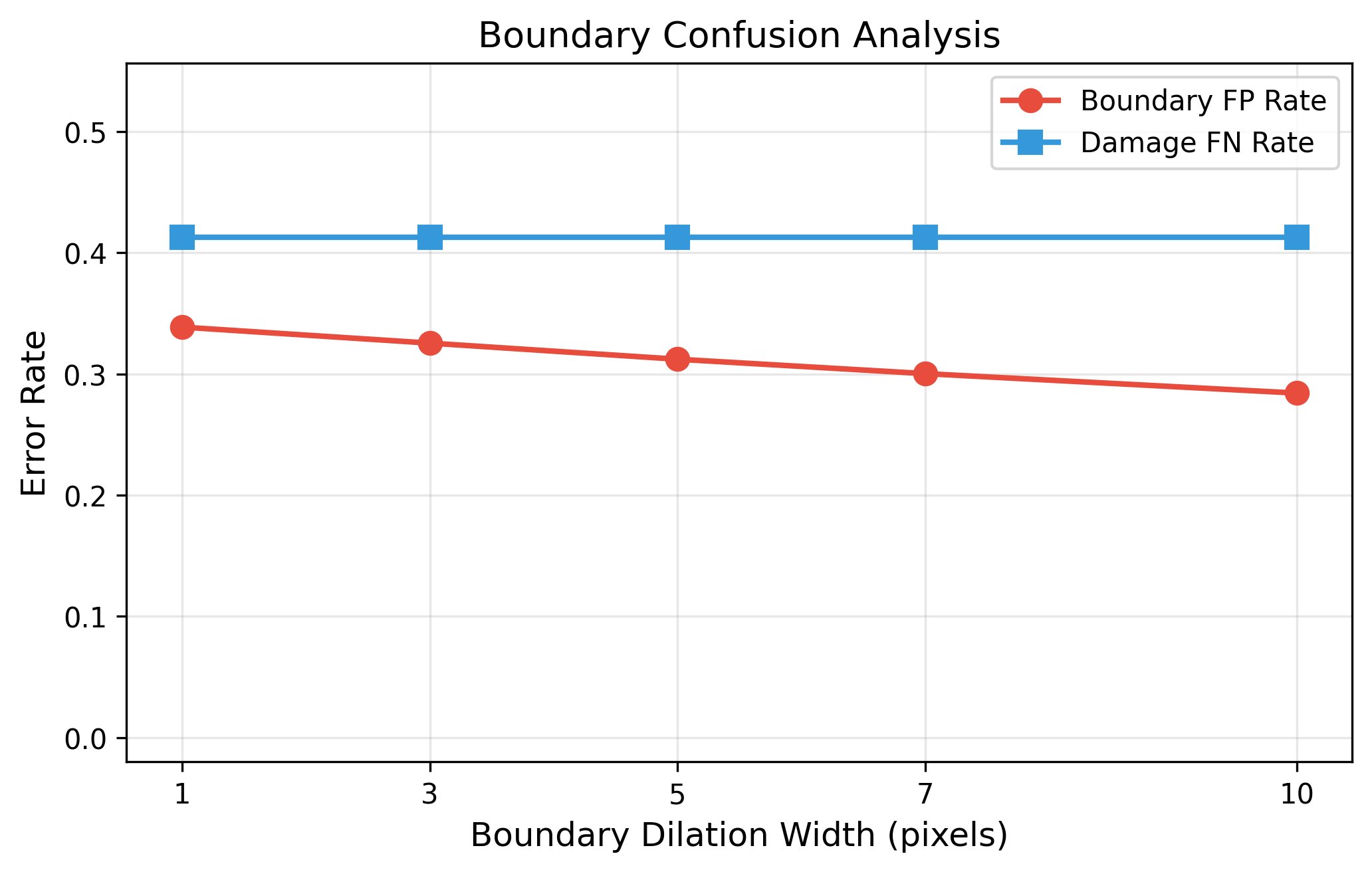}
        \caption{Boundary confusion}
    \end{subfigure}
    \hfill
    \begin{subfigure}[b]{0.48\linewidth}
        \includegraphics[width=\linewidth]{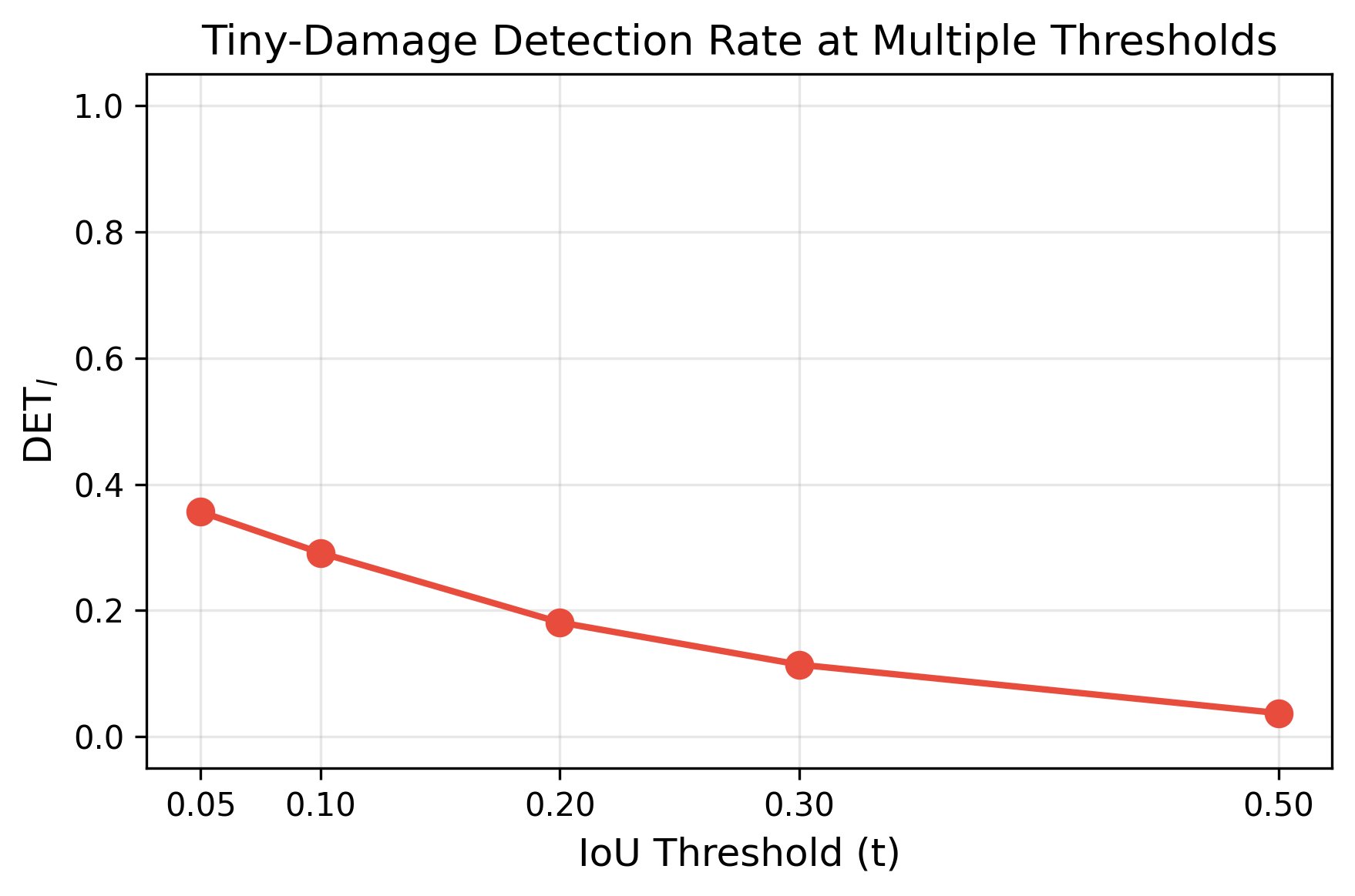}
        \caption{$\text{DET}_l$ at multiple thresholds}
    \end{subfigure}
    \caption{(a) Boundary false-positive rate decreases with wider analysis windows (0.34 at 1px to 0.29 at 10px) while damage false-negative rate stays flat at 0.41, confirming feature ambiguity (not boundary imprecision) as the core failure mode. (b) $\text{DET}_l$ drops from 0.35 ($t{=}0.05$) to 0.04 ($t{=}0.50$), showing the model finds the correct region with imprecise overlap.}
    \label{fig:analysis_plots}
\end{figure}

\subsection{Additional Qualitative Segmentation Results}

\begin{figure}[!htbp]
    \centering
    \begin{subfigure}[b]{0.19\linewidth}
        \includegraphics[width=\linewidth]{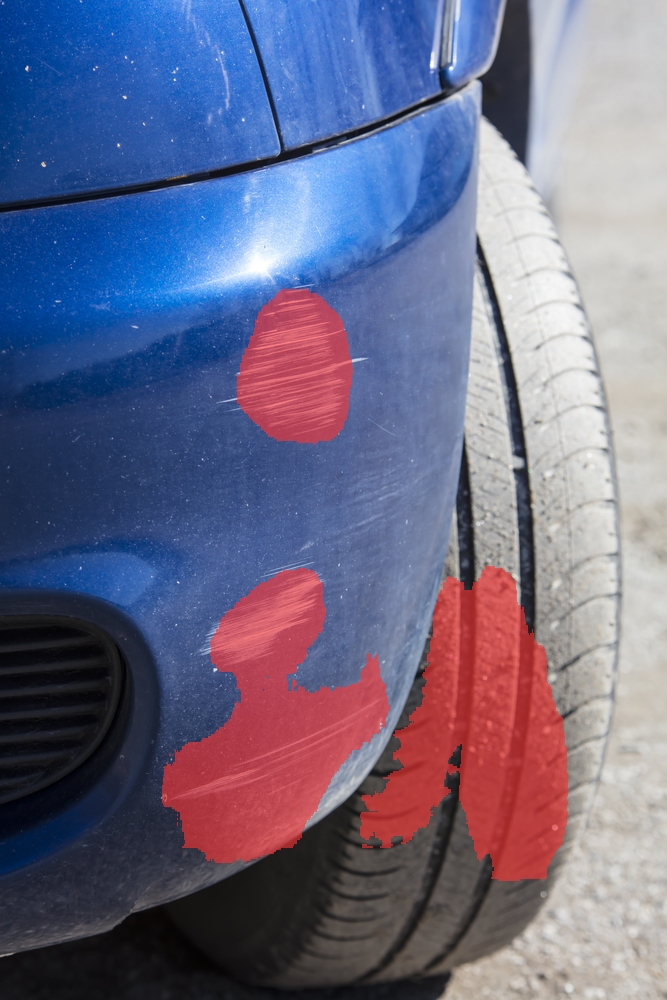}
        \caption{Scratch}
    \end{subfigure}
    \hfill
    \begin{subfigure}[b]{0.19\linewidth}
        \includegraphics[width=\linewidth]{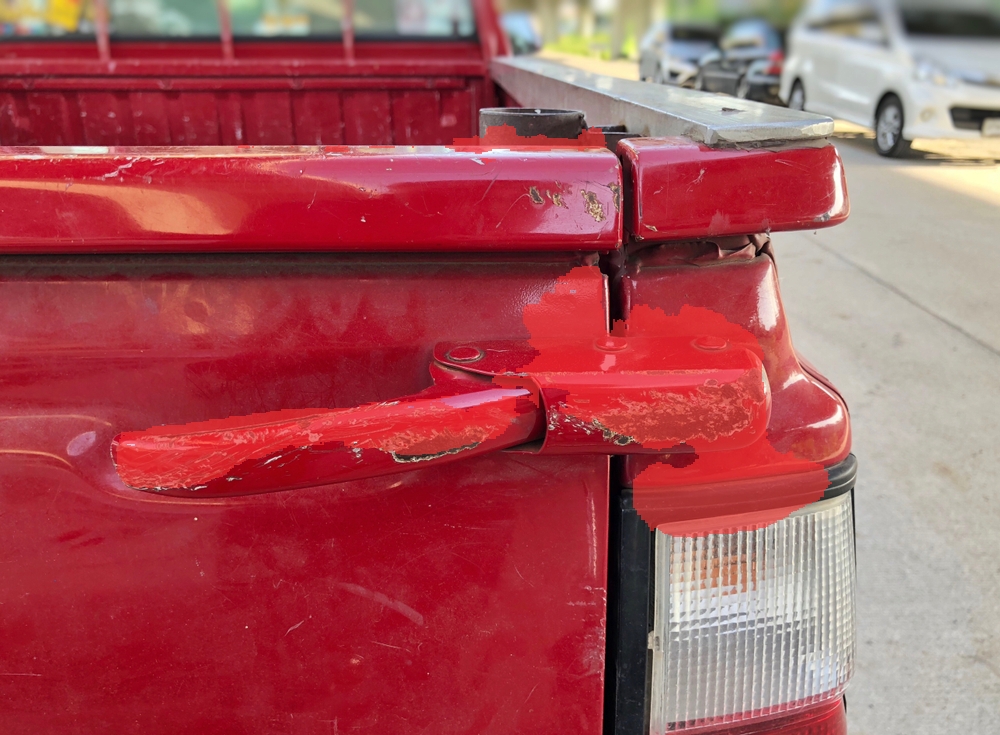}
        \caption{Scratch}
    \end{subfigure}
    \hfill
    \begin{subfigure}[b]{0.19\linewidth}
        \includegraphics[width=\linewidth]{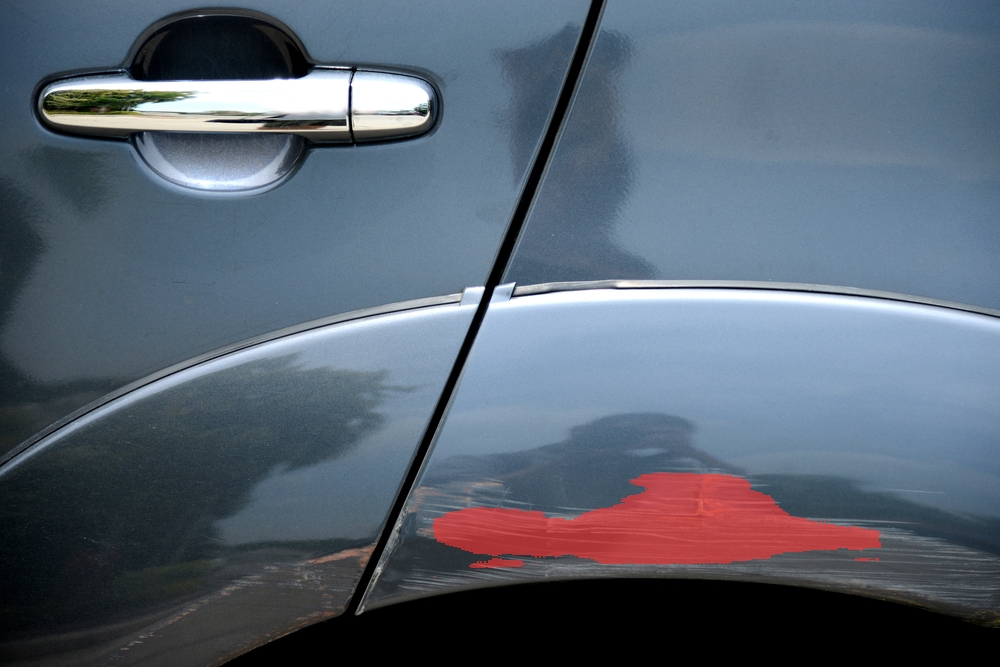}
        \caption{Dent}
    \end{subfigure}
    \hfill
    \begin{subfigure}[b]{0.19\linewidth}
        \includegraphics[width=\linewidth]{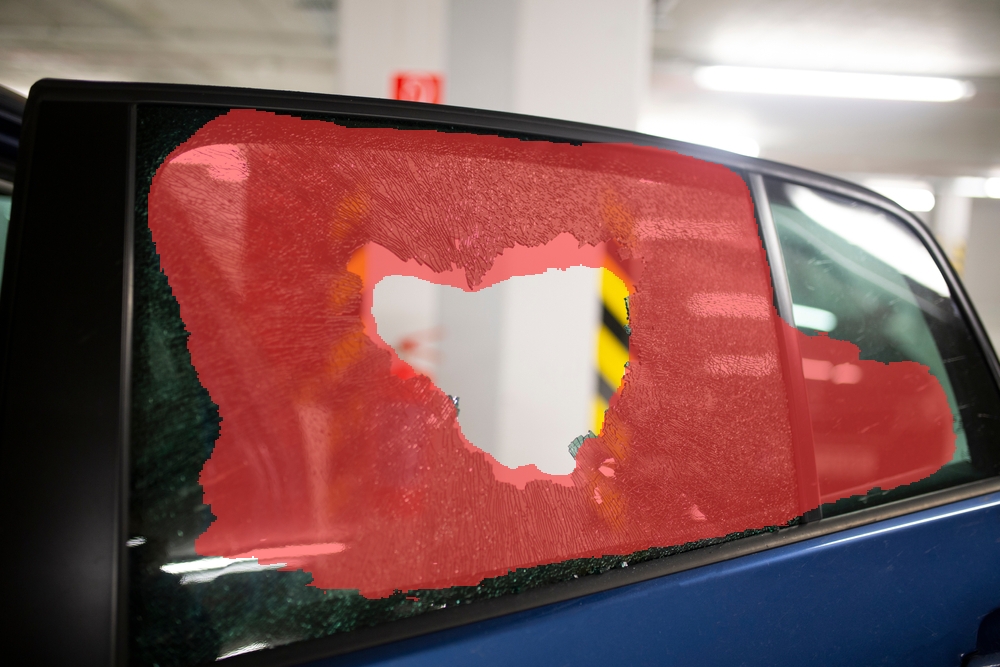}
        \caption{Glass}
    \end{subfigure}
    \hfill
    \begin{subfigure}[b]{0.19\linewidth}
        \includegraphics[width=\linewidth]{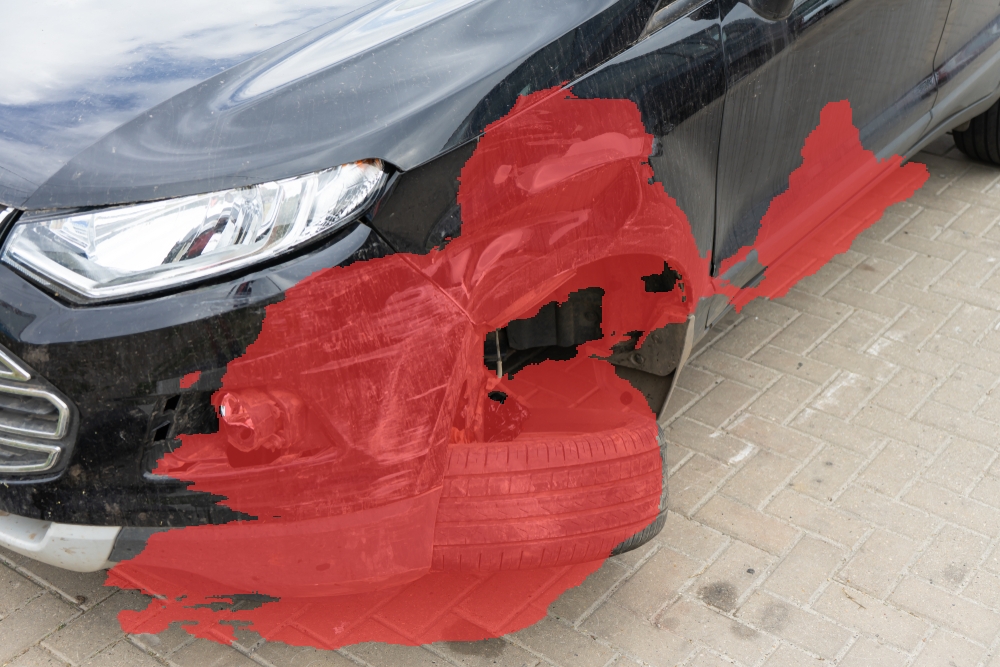}
        \caption{Crush}
    \end{subfigure}

    \vspace{3pt}

    \begin{subfigure}[b]{0.19\linewidth}
        \includegraphics[width=\linewidth]{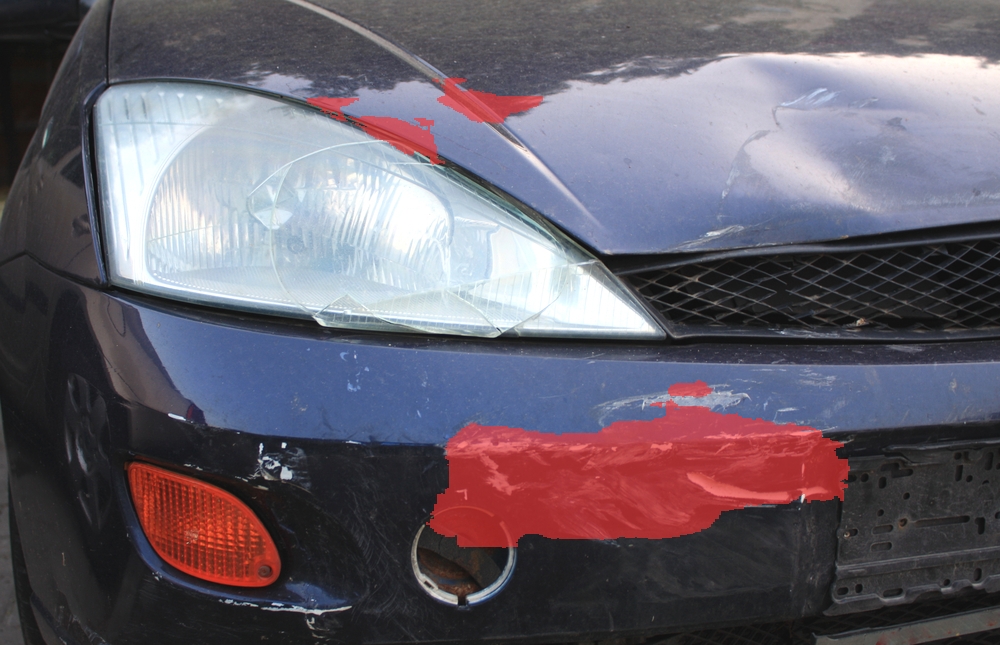}
        \caption{Scratch}
    \end{subfigure}
    \hfill
    \begin{subfigure}[b]{0.19\linewidth}
        \includegraphics[width=\linewidth]{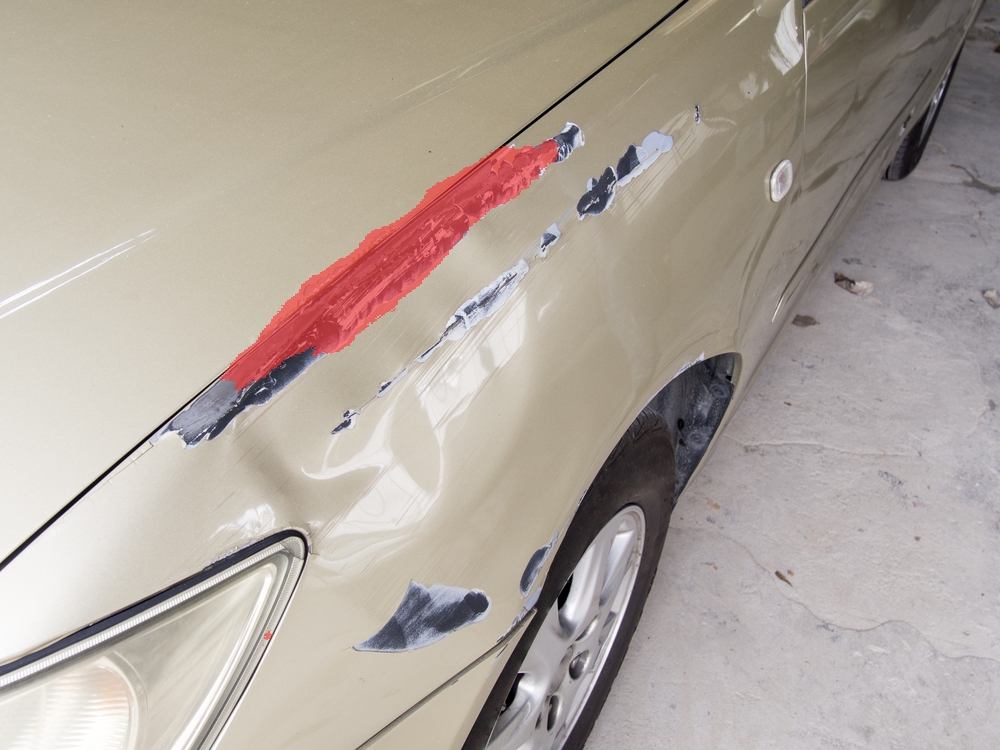}
        \caption{Scratch}
    \end{subfigure}
    \hfill
    \begin{subfigure}[b]{0.19\linewidth}
        \includegraphics[width=\linewidth]{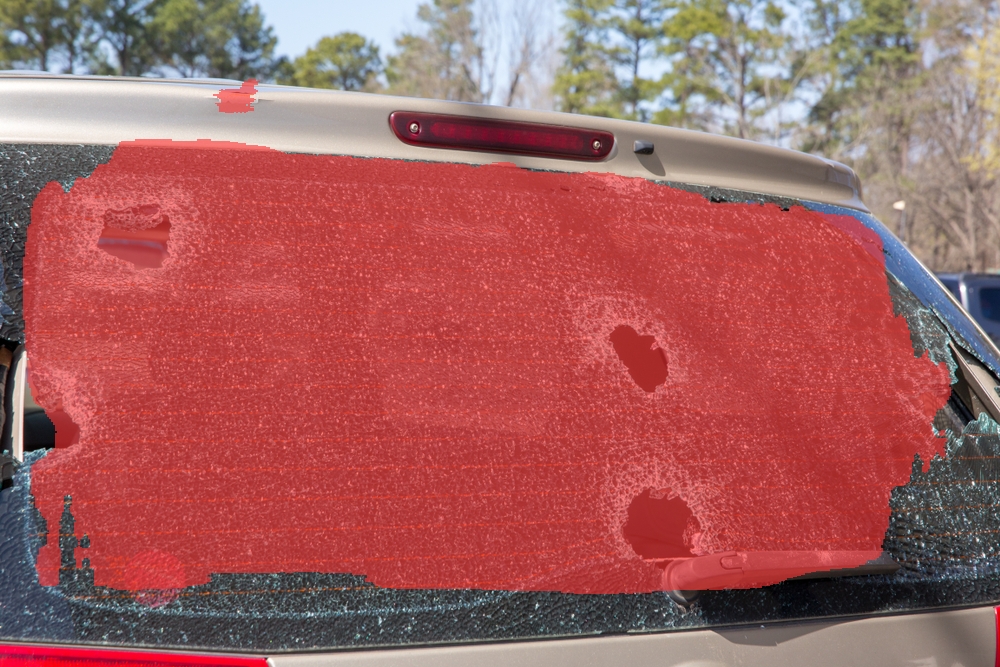}
        \caption{Glass}
    \end{subfigure}
    \caption{Additional TinyDamage inference results (red overlays) across CarDD damage categories, complementing Figure~\ref{fig:seg_results}.}
    \label{fig:qualitative}
\end{figure}

Figure~\ref{fig:qualitative} shows further TinyDamage inference examples beyond those in the main text (Figure~\ref{fig:seg_results}), spanning multiple instances per category.

\subsection{Auxiliary Damage Classification}

The auxiliary 6-class classification head (shared FPN encoder, $\sim$50K additional parameters, zero inference cost when unused) reaches 29.6\% accuracy / 0.355 macro F1 / 0.587 micro F1 at 30 epochs (random baseline: 16.7\%). The macro/micro F1 gap reflects CarDD's class imbalance: strong performance on frequent categories (scratch, dent), weaker on rare ones (crack, tire flat), mirroring the same imbalance challenge addressed by the segmentation model.

\subsection{Summary of All Approaches Explored}

\begin{table}[h]
\centering
\caption{Summary of all techniques explored, including unsuccessful directions, spanning perception (top), generation (middle), and infrastructure (bottom).}
\label{tab:approaches}
\small
\begin{tabular}{@{}p{3.2cm}p{2.8cm}p{1.0cm}p{5.5cm}@{}}
\toprule
\textbf{Approach} & \textbf{Category} & \textbf{Result} & \textbf{Key Takeaway} \\
\midrule
Focal loss (standalone) & Perception & Failed & Collapses tiny-damage detection to zero (DET$_l$=0.0) \\
Gradient boundary loss & Perception & Failed & Overwhelmed by large-damage edges; useless for tiny objects \\
CE+Dice+Cosine & Perception & Works & Strong baseline; cosine annealing critical for small objects \\
FPN multi-scale encoder & Perception & Works & Largest single improvement (+2.3\% mIoU) \\
Contrastive pixel embeddings & Perception & Works & Improves damage-background separability; small mIoU gain over FPN \\
6-class auxiliary classifier & Perception & Partial & 29.6\% acc (1.77$\times$ random); undertrained at 30 epochs \\
\midrule
Qwen-VL zero-shot classification & Grounding/Gen. & Works & 87.3\% accuracy without fine-tuning \\
Qwen-VL spatial localization & Grounding/Gen. & Failed & Hallucinates damage in reflections; misses thin scratches \\
Image-only prompting & Grounding/Gen. & Weak & VLM describes entire vehicle instead of damage regions \\
Text-only prompting & Grounding/Gen. & Weak & Correct but generic; lacks visual grounding \\
Image+text grounded prompting & Grounding/Gen. & Works & Best report quality (4.1/5 for damage reports) \\
Chain-of-thought (coverage) & Grounding/Gen. & Works & Prevents skipping damage categories in assessment \\
RAG for historical claims & Grounding/Gen. & Not done & No paired (image, claim text) data in CarDD \\
LoRA fine-tuning of Qwen-VL & Grounding/Gen. & Not done & No domain-specific training corpus; strong zero-shot baseline \\
\midrule
LangGraph 7-node StateGraph & Infrastructure & Works & Clean state management; node-level retry and debugging \\
LangFuse tracing & Infrastructure & Works & Full latency/token/error observability per node \\
Streamlit application & Infrastructure & Works & End-to-end interactive demo for claim generation \\
\bottomrule
\end{tabular}
\end{table}

\end{document}